\pdfoutput=1

\documentclass[11pt]{article}

\usepackage{acl}
\usepackage{hyperref}
\usepackage{booktabs}
\usepackage{times}
\usepackage{latexsym}
\usepackage{multirow}
\usepackage{amssymb}
\usepackage{siunitx}
\usepackage{xcolor}
\usepackage[many]{tcolorbox}

\usepackage{graphicx}
\usepackage{todonotes}
\usepackage{amsmath}
\usepackage{multirow}
\usepackage{multicol}
\usepackage{caption}
\usepackage{subcaption}

\usepackage[T1]{fontenc}

\usepackage[utf8]{inputenc}

\usepackage{microtype}

\usepackage{inconsolata}

\usepackage{graphicx}

\title{Understand the Implication: Learning to Think for\\ Pragmatic Understanding}

\author{Settaluri Lakshmi Sravanthi$^\ast$$^\dagger$,  Kishan Maharaj$^\ast$$^\dagger$,  Sravani Gunnu$^\dagger$,\\ \textbf{Abhijit Mishra}$^\ddagger$,  \textbf{Pushpak Bhattacharyya}$^\dagger$ \\
$\dagger$Indian Institute of Technology Bombay, Mumbai, India \\
$\ddagger$University of Texas at Austin, Texas, United States\\
   \texttt{\{sravanthi.settaluri, kishan.maharaj.iitb, sravi.gunnu\}@gmail.com,} \\
      \texttt{abhijitmishra@utexas.edu,
    pb@cse.iitb.ac.in}}

\begin{document}
\maketitle
\def\thefootnote{$\ast$}\footnotetext{Equal Contributions}\def\thefootnote{\arabic{footnote}}
\begin{abstract}
Pragmatics, the ability to infer meaning beyond literal interpretation, is crucial for social cognition and communication. While LLMs have been benchmarked for their pragmatic understanding, improving their performance remains underexplored. Existing methods rely on annotated labels but overlook the reasoning process humans naturally use to interpret implicit meaning. To bridge this gap, we introduce a novel pragmatic dataset \textbf{ImpliedMeaningPreference} that includes \textit{explicit reasoning (`thoughts')} for both correct and incorrect interpretations. Through preference-tuning and supervised fine-tuning, we demonstrate that thought-based learning significantly enhances LLMs' pragmatic understanding, improving accuracy by 11.12\% across model families. 
We further discuss a transfer-learning study where we evaluate the performance of \textit{thought}-based training for the other tasks of pragmatics (presupposition, deixis) that are not seen during the training time and observe an improvement of 16.10\% compared to \textit{label} trained models. Code and data are available in the repo \footnote{\href{https://github.com/SLSravanthi/Reasoning_and_Pragmatics.git}{Code and Datasets}} 

\end{abstract}

\section{Introduction}

Human interactions shape relationships through shared understandings, influenced not just by explicit words but by emotional and pragmatic nuances that convey implicit meanings. The ability to interpret beyond the literal meaning of language, known as \textit{pragmatics}, is essential for social cognition, interpersonal awareness, and emotional intelligence. It allows individuals to navigate conversations fluidly, recognising intentions, cultural contexts, and unspoken implications.

\begin{figure}[ht!]
    \centering
    \includegraphics[width=1\linewidth]{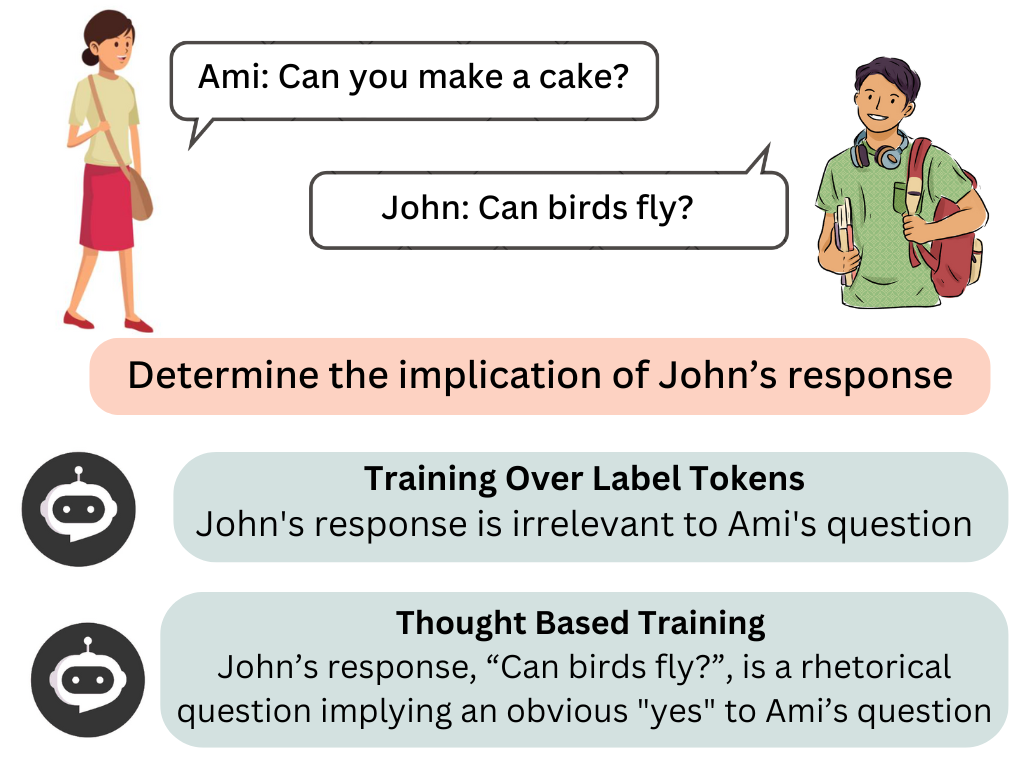}
    \caption{An example of implicature in a response which demonstrates the effectiveness of thought-based training by capturing the correct pragmatic meaning}
    \label{fig:implicature_example}
\end{figure}

Recent progress in large language models \cite{DBLP:conf/nips/BrownMRSKDNSSAA20, team2024gemma,qwen2.5,achiam2023gpt,dubey2024llama, team2023gemini} has advanced the capabilities of conversational AI. These systems exhibit robust performance in natural language generation, reasoning tasks like math word problems, code generation \cite{superglue,gsm8k,strategyqa,ARC}, etc., largely due to the exploitation of extensive computational resources and vast language datasets. Despite these strengths, current LLMs struggle with effective communication, specifically in capturing the pragmatic and ambiguous dimensions of user inputs. Additionally, conventional training strategies prioritise the production of responses that are safe, objective, and widely acceptable \cite{glaese2022improving}. This approach, while ensuring reliability, diverges from the goal of replicating truly human-like conversational behaviour, where the subtleties of context, emotion, and cultural nuance are critical.

While humans naturally engage in pragmatic reasoning, LLMs often struggle with this skill, especially the small LLMs (SLMs) \cite{amirizaniani2024can}, which are often used in practical scenarios due to their lower inference costs, reduced latency, and suitability for local deployment. Given the increased interaction between humans and LLMs, it is very important for the LLMs to obtain substantial pragmatic understanding of human language and intent. Recent work has primarily focused on evaluating LLMs' pragmatic understanding, yet efforts to enhance their performance on such tasks remain limited \cite{van2023large}.
\begin{figure*}[ht!]
    \centering
    \includegraphics[width=1\linewidth]{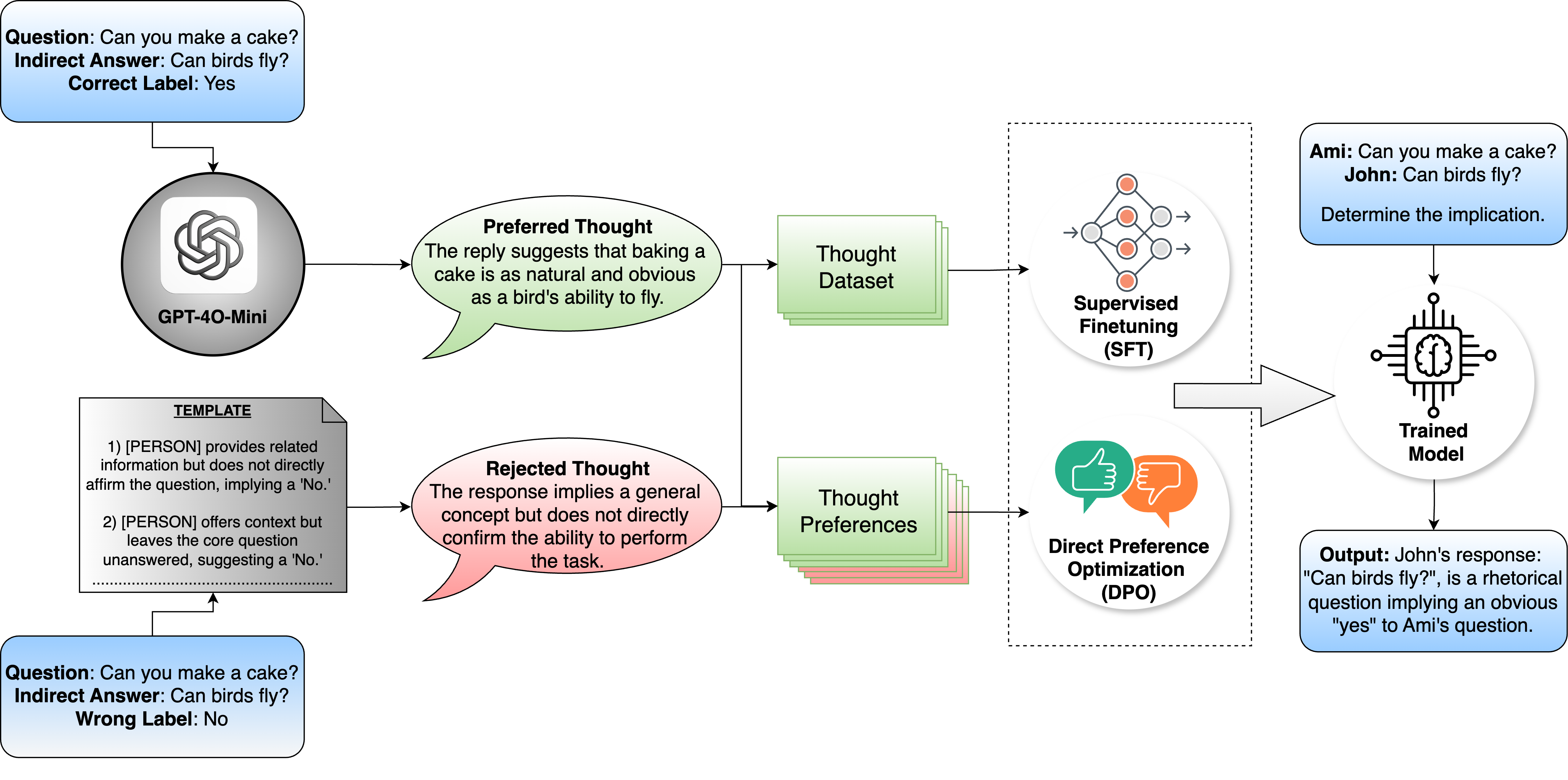}
    \caption{This diagram shows the proposed thought-based training framework with two different training mechanisms: 1) SFT (Supervised Finetuning) and 2) DPO (Direct Preference Optimisation). The left side of the diagram shows the preference data generation steps, and the right side of the diagram shows the training pipeline. We use preferred thought+label for SFT and preference tune with the rejected thought+incorrect label and preferred thought+correct label in DPO.}
    \label{fig:main_architecture}
\end{figure*}
Approaches that try to improve LLMs in pragmatic reasoning rely on label-based supervision or policy optimisation over annotated datasets \cite{wu2024rethinking}, but these methods do not explicitly incorporate the reasoning process that humans use to grasp implicit meaning. This is mainly due to the absence of training mechanisms which can explicitly incorporate the reasoning process. For instance, as shown in Figure \ref{fig:implicature_example}, interpreting the response \textit{"Can birds fly?"} as \textit{"Yes"} to the question \textit{"Can you make a cake?"} requires recognising it as a rhetorical question with an obvious affirmative answer—implying that the speaker’s answer to the original question is also an obvious \textit{"Yes"}. 

To address this gap, we introduce a novel approach that leverages explicit reasoning, or \textit{thoughts}, to improve LLMs' pragmatic comprehension. Specifically, we perform thought-based training for the task of implicature recovery, understanding what is implied in a statement even though it is not literally expressed. We then show generalizability on multiple pragmatics domains, which include implicature, presupposition and reference. Unlike reasoning tasks such as math word problems or coding challenges, pragmatic reasoning often lacks definitive answers, making it more challenging. The correct interpretation in a given scenario is highly influenced by context, culture, and the individuals involved. This interpretation is often not described in the raw training data explicitly and can not be easily captured during the training process. To mitigate this, an explicit intermediate reasoning process must be provided during the training time along with the correct label, which details the intermediate reasoning process, mimicking how humans derive correct interpretation by deliberate system-2 thinking \cite{weston2023system}. Hence, we present a first-of-its-kind pragmatic dataset where each instance includes a \textit{thought} explaining the reasoning behind the correct label, along with a plausible yet incorrect \textit{negative thought} justifying the incorrect label. We integrate this thought-based data into both preference-tuning and supervised fine-tuning settings, demonstrating an absolute improvement of \textbf{11.12\%} in accuracy across three model families. Our findings establish the effectiveness of thought-based learning in advancing LLMs' ability to interpret implicit meaning in language. Our contributions are:

\begin{itemize}

   \item A training framework incorporating explicit reasoning (\textit{thoughts}) \footnote{Here, thoughts do not imply human cognition.}, leading to an  11.12\% improvement in implicature recovery compared to label-based training approaches (Figure \ref{fig:main_architecture}).
    
    \item A transfer learning analysis examining the effects of thought-based supervised fine-tuning (SFT) and direct preference optimisation (DPO) on unseen tasks, showing an improvement of 16.10\% over label-based training approaches (Section \ref{sec: transfer learning}).
    \item Synthetic QA datasets; \textbf{Syn-Circa }and \textbf{Syn-ludwig}, consisting of $\sim$33.75K, created by extending CIRCA and LUDWIG to improve understanding of implicit responses (Section \ref{subsec: synthetic-data}). 
    
    \item A novel dataset, named \textbf{ImpliedMeaningPreference}, for thought based implicature recovery consisting of \textit{$\sim$66.2K} instances. This dataset is developed through a human-LLM collaboration integrating multiple implicature recovery datasets (Section \ref{subsec:preference data}).
    

\end{itemize}

\section{Related Work}
Implicature recovery is a central topic in pragmatics, attracting significant attention from linguists and computational researchers alike. One of the most influential theoretical contributions to this field is the formulation of the 	\textit{Gricean Maxims} \citep{grice1975logic}, which outline principles governing conversational implicature through Quality, Quantity, Relevance, and Manner.

Various approaches have been proposed to analyse and recover implicatures. For instance, \citet{circa,ruis2023goldilocks} study indirect answers in polar questions, shedding light on how conversational participants infer unstated meanings. \citet{grice} leverage hierarchical grammar models to interpret both implicatures and deictic references in structured dialogues. Additionally, \citet{impres} explores the role of Natural Language Inference (NLI) in understanding scalar implicatures, while \citet{Implicature} integrate implicature-based reasoning into sentiment analysis. 

Further contributions in this domain include corpus-based studies such as \citet{squinky}, which provide sentence-level annotations for implicature detection. Work by \citet{schuster2019harnessing} and \citet{li2021predicting} focuses on employing neural networks and linguistic signals to predict scalar inferences, highlighting the potential of machine learning in implicature comprehension. Despite these advancements, recent benchmarking efforts \citep{hp,sravanthi2024pub} consistently reveal a persistent performance gap between human reasoning and LLM capabilities in pragmatics.

Building upon these findings, \citet{wu2024rethinking} introduces an open-ended evaluation framework to assess LLMs' pragmatic abilities, showing the superiority of preference-based learning over supervised fine-tuning when label-based data is considered. Going forward, our work incorporates the intermediate reasoning steps (thoughts) in fine-tuning and preference optimisation process for pragmatic reasoning. Unlike conventional approaches that reward only label accuracy, our method explicitly incorporates thought processes into model training, enabling LLMs to develop a deeper understanding of pragmatics. In the following sections, we present our datasets, methodology and evaluate the effectiveness of structured reasoning in enhancing LLM performance in pragmatics tasks like implicature recovery, presupposition and deixis.

\section{Datasets}
In this section, we discuss the process of generating ImpliedMeaningPreference data and synthetic QA datasets (Syn-Circa and Syn-ludwig). 
\subsection{Preference Data Generation}
\label{subsec:preference data}
Gathering high-quality preference data typically requires substantial resources and significant human effort. Existing pragmatic QA datasets, such as Circa and Ludwig, include human-annotated mappings between indirect answers and their corresponding direct interpretations (i.e., labels such as yes and no). To minimise human efforts in preference construction, we leverage these existing label mappings: the original mapped label is treated as the preferred label, while its complement is considered the \textit{rejected label}. \\
\textbf{Preferred thought generation:} As shown in Figure \ref{fig:main_architecture} we generate the \textit{thoughts} supporting the \textit{correct label} by prompting \textit{gpt-4o-mini}. \textit{<Question, Indirect answer, Correct label>} are given as input to the model, and the model is tasked to generate an intermediate reasoning step that helps in mapping the indirect answer to the label. \\
\textbf{Rejected thought generation:} We attempted to generate rejected thoughts using a similar approach by providing \textit{<Question, Indirect answer, Wrong label>} to the model. However, we observed that most of the time, the model generated thoughts supporting the \textit{correct label}, which may be due to the inherent safety guardrails present in the model \cite{achiam2023gpt}. Therefore, for rejected thought generation, a linguistic expert is tasked to write templates. The templates are made to capture the wrong reasoning that mimics the misunderstanding humans can have when one or more of the Gricean maxims are flouted. For example, in the Figure \ref{fig:implicature_example}, the \textit{maxim of relevance} is flouted, and it can be understood as \textit{John} is giving an irrelevant reply to the question asked by \textit{Ami}. Each \textit{rejected thought} is generated by randomly selecting one of the 50 templates written by the linguist. Prompts for the data generation and sample templates are given in the Appendix, Section \ref{sec: transfer learning}. A sample of preferred and rejected thoughts is verified by linguistic experts. Details discussing the human evaluation can be found in Appendix \ref{appendix: thought_evaluation}.

\subsection{Synthetic QA Datasets}
\label{subsec: synthetic-data}
To enhance our preference dataset, we expanded the existing QA dataset, facilitating the generation of additional preference annotations. We construct our synthetic QA datasets based on existing polar questions and indirect answer datasets \cite{circa,ruis2023goldilocks}. 
 Circa \cite{circa} and ludwig \cite{ruis2023goldilocks} consist of 3,345 and 601 unique questions. For generating \textit{syn-circa} and \textit{syn-ludwig}, we take unique questions from both the datasets and generate indirect answers that can be mapped to polar direct answers \textit{i.e.}, each indirect answer conveys a "yes" or "no" reply to the question. 
 The responses were generated using \textit{gpt-4o-mini} \cite{achiam2023gpt} in a few-shot prompting setting. For each unique question, we generate five answers using five different temperature values -0.0, 0.2, 0.4, 0.6 and 0.8 for generating varied and creative responses. To guide the model effectively, 3 to 6 examples were randomly selected from a curated set of 50 examples, serving as contextual prompts to steer the generation process. The prompts for generation are given in Appendix \ref{appendix: generation_prompt}. To evaluate that the generated responses adhered to the desired criteria of being indirect, we use a BERT-based classifier \cite{devlin2018bert} trained on classifying declarative sentences of the questions and indirect answers. Out of 33.75K instance only five examples were classified as indirect answers by the classifier. The effects of this data augmentation using synthetic datasets is discussed in Section \ref{data_ablation}.

\begin{table*}[ht]
\centering
\begin{tabular}{@{}|l|c|c|c|c|c|c|c|c|c|@{}}
\toprule
\textbf{Dataset Name}  & \multicolumn{3}{|c|}{\textbf{Train Set}} & \multicolumn{3}{|c|}{\textbf{Val Set}} & \multicolumn{3}{|c|}{\textbf{Test Set}} \\ \hline
 & \textbf{Yes} & \textbf{No} & \textbf{Total} & \textbf{Yes} & \textbf{No} & \textbf{Total} & \textbf{Yes} & \textbf{No} & \textbf{Total} \\ \midrule
Circa                 & 11,996       & 9,310       & 21,306         & 2,957        & 2,251       & 5,208          & 1,675        & 1,272       & 2,947          \\
Synthetic\_Circa      & 9,955        & 9,517       & 19,472         & 2,502        & 2,468       & 4,970          & 1,322        & 1,394       & 2,716          \\
Synthetic\_Ludwig     & 2,401        & 2,330       & 4,731          & 592          & 608         & 1,200          & 327          & 333         & 660            \\ \bottomrule
\end{tabular}
\caption{Class distribution and totals for Train, Validation, and Test datasets.}
\label{tab:dataset_distribution}
\end{table*}

\section{Methodology}
This section discusses our approach in detail. We aim to study the impact of incorporating thought training in two settings: 1) Supervised Fine-Tuning (SFT) and 2) Direct Preference Optimization (DPO) \cite{rafailov2023direct}. We formulate the task as follows: Given an input consisting of an indirect answer to a question along with a context, output the pragmatic interpretation. Let $P(x)$ be the initial prompt which contains the task description $T_{desc}$ and input description $I_{desc}$ where $x=[T_{desc}, I_{desc}]$. Let $G$ be the generated output, which contains a thinking process $\mathcal{T}_{thought}$ followed by a predicted label $\mathcal{P}_{label}$. Here, $G=[\mathcal{T}_{thought}; \mathcal{P}_{label}]$ consisting of tokens $( g_1, g_2, \dots, g_{t-1}, g_t)$.     

In the general supervised fine-tuning process, we aim to maximize the conditional log-likelihood of the output tokens given the input tokens. In the context of our setting, this corresponds to:

\[
\mathcal{L}_{sft} = -\sum_{t=1}^{|G|} logP_{\theta}(g_t \mid P(x), g_1, g_2, \dots, g_{t-1})
\]

Here \( \mathcal{L} \) is the total loss (negative log-likelihood of the sequence),
\( |G| \) is the length of the output sequence \( G \), \( P_{\theta}(g_t \mid P(x), g_1, g_2, \dots, g_{t-1}) \) is the model’s predicted probability of the token \( g_t \) at position \( t \), given the input prompt \( P(x) \) and all previous tokens \( g_1, g_2, \dots, g_{t-1} \) and \( \theta \) is the model parameters being optimized.

Contrary to SFT, in standard Reinforcement Learning from Human Feedback (RLHF) setup, we use the structure of the Markov Decision Process consisting of 4 tuples: (States $S$, Actions $A$, Transition Probabilities $T_p$, Rewards $R$). Here, we define a function policy $\pi$, which maps states to actions ($\pi:S \rightarrow A $). The goal is to optimize the policies to maximize the rewards. In our context, given the current state (Input Prompt), we would like to optimize the policy (language model) to select the actions (which token to predict next) such that the reward function (a function which scores the generated output based on human preferences) yields the maximum value. We aim to study the effects of optimizing policy over the thoughts and labels together. 

This means that the probability of winning generation ($G_W$) preferred by humans should be more than the probability of losing generation ($G_L$), which humans do not prefer. Therefore, the Bradley Terry Model for our setup is: 

\begin{equation}
P(G_W>G_L)=\frac{e^{R(P(x), G_W)}}{e^{R(P(x), G_W)}+e^{R(P(x), G_L)}}
\end{equation}

This finally yields the adapted DPO loss for our setting, incorporating policy optimization over thought and labels.

Specifically, $L_{DPO}(\pi_{\theta}; \pi_{ref})$:
\begin{equation}
   -\mathbb{E}_{(x, G_W, G_L)\sim\mathbb{D}}[log(\sigma(\beta\psi(G_W) -\beta\psi(G_L))]
\end{equation}

where 

\begin{equation}
    \psi(G)=log(\frac{\pi_{\theta}(G|P(x))}{\pi_{ref}(G|P(x))})
\end{equation}

In the above equation, $\pi_{ref}$ is the reference model instantiated with the initial version of the model, $\pi_{\theta}$ is the model obtained after preference tuning, and $\beta$ is the regularizing parameter used for penalizing the scenario when the resulting model is very far from the base version resulting to the loss of prior knowledge.  

From a linguistic perspective, our approach is motivated by the need to model pragmatic competence in language understanding. Pragmatic reasoning involves interpreting implied meanings that go beyond the literal content of utterances, as theorized in Grice’s maxims of conversation. Traditional models often struggle with implicature resolution because they lack an explicit mechanism for reasoning about contextually inferred meanings. By integrating structured thought processes into both fine-tuning and preference optimization, our method provides a computational analog to human inferential processes in discourse interpretation. This, we hope, should enable LLMs to better grasp implicatures, handle indirect responses, and align with human-like conversational norms, thereby improving their effectiveness in pragmatic language tasks.

\section{Experimental Setup}
For our experiments, we consider models from three different families: 1) Llama-3.2-1B \cite{dubey2024llama} 2) Qwen-2.5-1.5B \cite{yang2024qwen2}, 3) Gemma-2-2B \cite{team2024gemma}. We also report the zero-shot performance of Llama3.1 70B for comparison with a large language model. Our experiments kept the learning rate at $5e-7$ with warmup steps of 500 iterations. We use RMSprop \cite{ruder2016overview} as our optimiser following \citet{wu2024rethinking}. All models in both settings are trained for one epoch (till convergence), and the greedy decoding mechanism was used throughout the experiments. For Qwen-2.5-1.5B and Gemma-2-2B, we use the global batch size of 32; for llama-3.2-1B, the global batch size was set to 64. For regularization, we use gradient clipping of 1 in DPO and weight decay of 0.01 in SFT. We use 4 NVIDIA H100 80GB HBM3 GPUs for all the experiments in this work, with a total train time of 8 GPU hours. For all other hyper-parameters, we use the default values. 
We report all the training and evaluation prompts in Appendix, Sections \ref{Appendix: Train prompts} and \ref{Appendix: Eval prompts} respectively. We use macro precision (P), recall (R) and F1 scores for evaluation.


\section{Results}
In Table \ref{tab:QA_data_comparison}, we report the results of our experiments after training with QA datasets. We note significant improvement after the inclusion of the thoughts in SFT and DPO for Llama-3.2-1B and Qwen2.5-1.5B. For Gemma2-2B, we observe significant gains in SFT with thought settings and a slight performance decline when thought is incorporated in the DPO setting. We note that DPO was not originally used in the training process of Gemma-2B, unlike Llama-3.2-1B and Qwen2.5-1.5B. We conjecture that since the model was not exposed to DPO in the general training, our training for implicature recovery could not induce the thoughts as effectively as in other models.

We note that training with just labels provides an edge to DPO over SFT across models, aligning with \citet{wu2024rethinking}. While training with thought alongside labels provides significantly higher gains for SFT when compared to DPO, with thought-based SFT outperforming thought-based DPO in most cases.  
The `thoughts' contain more explicit signals and the interpretation of the reasoning required to reach the right answer, which may be captured in a more straightforward way in the SFT setup compared to DPO. Intuitively, the thought-based training mechanism would require higher updates in the parameters than the scenario when we have to optimise over only label tokens. In general, the optimisation objective for SFT does not have any constraints and is more flexible compared to DPO, which requires a regularising parameter $\beta$ for the KL constraint to prevent divergence from the base model (untrained) during the training.

\begin{table*}[ht!]
\centering
\resizebox{\textwidth}{!}{
\begin{tabular}{|l|ccc|ccc|ccc|c|}
\hline
\textbf{Setting}         & \multicolumn{3}{c|}{\textbf{Circa}} & \multicolumn{3}{c|}{\textbf{Synthetic\_Circa}} & \multicolumn{3}{c|}{\textbf{Synthetic\_Ludwig}} & \textbf{Overall F1} \\ \hline
 & \textbf{P} & \textbf{R} & \textbf{F1} & \textbf{P} & \textbf{R} & \textbf{F1} & \textbf{P} & \textbf{R} & \textbf{F1} &  \\ \hline

\multicolumn{11}{c}{\textbf{Llama-3.2 1B}} \\ \hline
\textbf{Zero-Shot}       & 73.52           & 60.67           & 50.83            & 79.35           & 62.90           & 57.49            & 79.38           & 65.60           & 61.24            & 56.52 \\ \hline
\textbf{SFT}             & 68.28           & 55.57           & 42.72            & 72.53           & 53.84           & 42.39            & 64.68           & 56.32           & 49.27            & 44.79 \\ 
\textbf{SFT+Thought}     & 81.32           & 81.87           & 81.40 \textcolor{teal}{(+38.68)}           & 86.37           & 86.39           & 86.37 \textcolor{teal}{(+43.98)}            & 82.01           & 79.42           & 79.09 \textcolor{teal}{(+29.82)}           & 82.29  \textcolor{teal}{(+37.5)} \\ \hline
\textbf{DPO}             & 81.17           & 60.22           & 55.37            & 77.67           & 61.73           & 54.62            & 94.14           & 93.83           & 93.78            & 67.92 \\ 
\textbf{DPO+Thought}     & 75.39           & 73.81           & 74.16 \textcolor{teal}{(+18.79)}            & 81.77           & 80.95           & 80.62 \textcolor{teal}{(+26)}             & 89.01           & 88.82           & 88.78 \textcolor{red}{(-5)}             & 81.85 \textcolor{teal}{(+13.93)}  \\ \hline

\multicolumn{11}{c}{\textbf{Qwen-2.5-1.5B}} \\ \hline
\textbf{Zero-Shot}       & 75.15           & 63.15           & 54.55            & 83.51           & 75.78           & 74.78            & 86.73           & 81.96           & 81.49            & 70.94 \\ \hline
\textbf{SFT}             & 74.39           & 62.53           & 53.76            & 85.04           & 79.23           & 78.75            & 87.00           & 82.11           & 81.63            & 71.38 \\ 
\textbf{SFT+Thought}     & 76.85           & 70.13           & 64.99 \textcolor{teal}{(+11.23)}             & 87.90           & 84.30           & 84.24 \textcolor{teal}{(+5.49)}            & 88.72           & 86.10           & 85.96  \textcolor{teal}{(+4.33)}           & 78.40 \textcolor{teal}{(+7.02)} \\ \hline
\textbf{DPO}             & 76.96           & 72.49           & 68.50            & 88.10           & 87.23           & 87.31            & 88.77           & 86.72           & 86.63            & 80.81 \\ 
\textbf{DPO+Thought}     & 77.77           & 74.93           & 71.80 \textcolor{teal}{(+3.3)}            & 90.13           & 89.02           & 89.11 \textcolor{teal}{(+1.8)}           & 88.22           & 87.52           & 87.51 \textcolor{teal}{(+0.88)}            & 82.14 \textcolor{teal}{(+1.33)} \\ \hline

\multicolumn{11}{c}{\textbf{Gemma2-2B}} \\ \hline
\textbf{Zero-Shot}       & 77.52 &   71.94    &67.46             & 79.18  & 67.82   & 64.91            & 83.93 &   77.38 &   76.39            & 69.58 \\ \hline
\textbf{SFT}             & 83.25           & 81.60           & 79.15            & 93.42           & 92.26           & 92.38            & 84.75           & 78.14           & 77.20            & 82.24 \\ 
\textbf{SFT+Thought}     & 90.69 &   91.10  &  \textbf{90.85} \textcolor{teal}{(+11.7)}           & 95.61  &  95.63  &  \textbf{95.58} \textcolor{teal}{(+3.2)}      & 93.48 &   93.49  &  \textbf{93.48} \textcolor{teal}{(+16.28)}            & \textbf{93.30} \textcolor{teal}{(+11.06)}\\ \hline
\textbf{DPO}             & 87.55           & 82.88           & 83.77            & 89.45           & 87.61           & 87.18            & 94.57           & 94.54           & 94.54            & 88.50 \\ 
\textbf{DPO+Thought}     & 87.48           & 80.63           & 81.54 \textcolor{red}{(-2.33)}            & 84.75           & 80.00           & 78.87 \textcolor{red}{(-8.31)}            & 92.59           & 92.17           & 92.10 \textcolor{red}{(-2.44)}           & 84.17 \textcolor{red}{(-4.33)}\\ 
\hline
\multicolumn{11}{c}{\textbf{Llama3.1-70B}}\\
\hline
\textbf{Zero-Shot}      &  94.37    & 93.88   & 94.09 & 93.61  &  93.63   & 93.59 & 91.79   & 91.69   & 91.66 &  93.11 \\ 
\hline

\end{tabular}
}
\caption{Comparison of P (Precision), R (Recall), and F1 scores across Circa, Synthetic\_Circa, and Synthetic\_Ludwig datasets under various settings for QA dataset. The last column reports the mean F1 score across datasets.}
\label{tab:QA_data_comparison}
\end{table*}

Another perspective in the context of this observation was suggested by \citealp{feng2024towards, pal2024smaug}, which shows the gradient of the DPO loss with respect to preferred (winning) response is lower compared to the dispreferred (losing) response which essentially hinders the learning capacity of LLMs to generate the actual human preferences while introducing the tendency of avoiding human dispreferred responses. This effect may have been further magnified in our setting which has more tokens compared to the only label setting.    

We also note that our best-performing model, Gemma2-2B, supervised-fine-tuned with thoughts, yields comparable performance to LLama3.1-70B, which highlights the effectiveness of incorporating thoughts in the training mechanism. In general, the thought-based training mechanism yielded better results compared to the setting, which just incorporates labels, highlighting the importance of learning thought generation.

\section{Analysis}
In this section, we discuss various insights about the proposed method, which describes the advantages of thought-based learning and some general error cases.

\subsection{Predictive Analysis}
Here, we describe the general predictive trends observed in our framework. 
In general, we observe a significant improvement after incorporating thought in the generated output. Intuitively, the causal models are optimized to generate the appropriate explanations first and then derive the predictions based on the generated explanation. Probabilistically, the next token for prediction is conditioned on the `thought' and `input' tokens which can act as a guide for reaching the correct predictions more accurately compared to the scenario when only input tokens are considered. We discuss an example in 
\hyperref[box: predictive_analysis_qa]{Illustration 1} where the task is to determine if the given response to a question implies a ``Yes" or `No". We observe that the model is correctly predicting the output of `Yes' by generating thought which are used for resolving the final predictions compared to the scenario when the model trained on just labels (without thought) is considered. The generated thought is also helpful in understanding if the model is genuinely predicting the correct output based on the right understanding or predicting the correct output randomly (further explanations in Section \ref{subsec:thought_analysis}).  

\begin{tcolorbox}[,
    colback=gray!10,     
    colframe=gray!30,    
    coltitle=black,
    fonttitle=\bfseries, 
    top=0mm,
    left=0mm,            
    right=0mm,           
    segmentation style={solid, black!80},
    bottom=-1mm-0.5mm,
    float=ht!
]
\small

\textbf{Given Input}: Determine whether a given response to a question implies a "Yes" or "No"\\
Context: Y has just told X that he/she is thinking of buying a flat in New York.\\
Question: Is it crowded and busy in that area?\\
Response: It never sleeps.
\tcbline

\textbf{With Thought Prediction}:
The response "It never sleeps" suggests that the area is always active and bustling with people, implying high population density and activity levels. This strongly indicates that the area would indeed be crowded and busy.\\
Prediction: Yes

\textbf{Without Thought Prediction}: No      \\
\textbf{Correct Label}: Yes

\tcbsubtitle[halign=center] {Illustration 1: General Prediction for QA}

\end{tcolorbox} 
\label{box: predictive_analysis_qa}

\begin{table*}[htbp]
\centering
\resizebox{\textwidth}{!}{
\begin{tabular}{|l|ccc|ccc|ccc|c|}
\hline
\textbf{Setting} & \multicolumn{3}{c|}{\textbf{Circa (\%)}} & \multicolumn{3}{c|}{\textbf{Synthetic\_Circa (\%)}} & \multicolumn{3}{c|}{\textbf{Synthetic\_Ludwig (\%)}} & \textbf{Mean F1 (\%)} \\
\hline
 & \textbf{P} & \textbf{R} & \textbf{F1} & \textbf{P} & \textbf{R} & \textbf{F1} & \textbf{P} & \textbf{R} & \textbf{F1} & \\
\hline
\textbf{DPO} & 71.43 & 67.05 & 66.91 & 78.93 & 78.56 & 78.34 & 73.58 & 72.96 & 72.72 & 72.66 \\
\textbf{DPO+Thought} & 74.42 & 53.36 & 43.61 & 71.10 & 52.68 & 38.83 & 75.27 & 51.95 & 37.33 & 39.92 \\
\hline
\textbf{SFT} & 17.13 & 16.63 & 16.86 & 12.40 & 12.44 & 12.40 & 31.30 & 31.40 & 31.29 & 20.18 \\
\textbf{SFT+Thought} & 80.12 & 79.21 & 77.12 & 87.33 & 86.00 & 86.07 & 89.45 & 89.41 & 89.39 & 84.19 \\
\hline
\end{tabular}
}
\caption{\textbf{Data Ablation on Gemma-2B}: We report the Precision (P), Recall (R) and F1 scores on all four settings by training the model with just the Circa dataset.}
\label{tab:data ablation}
\end{table*}

\subsection{Transfer Learning Analysis}
\label{sec: transfer learning}

This section discusses whether thought learning is transferable to the other datasets and tasks which are not seen during the training process. The primary motivation behind this study is to understand if the thought training done for one of the pragmatic tasks is helpful in learning other pragmatic tasks in different datasets.  

Specifically, we evaluate our models trained for implied question answering with the following datasets: 
1) FigQA (figurative Natural Language Inference) 2) Flute (figurative Natural Language Inference) 3) IMPPRES (figurative Natural Language Inference) 4) Ludwig
5) \textit{Pub-presupposition} task 6)  \textit{Pub-reference} task.

Presupposition, implicature, and reference are pragmatic phenomena that rely on context, shared background knowledge, and the interactive nature of communication to convey meaning beyond the literal content of an utterance. Intuitively, models trained with explicit reasoning for performing implicature recovery should also perform better on these related linguistic phenomena. Specially, reference is a special case of implicature where the only difference is the usage of \textit{deixis} terms. 

For these experiments, we chose our best-performing model, Gemma2-2B. We observe significant improvement in performance when the thought is incorporated into the training mechanism for both the SFT and DPO. We report these results in Table \ref{tab: transfer_learning_nli} for three NLI datasets and Table \ref{tab: transfer_learning_others} for other pragmatics tasks with a mean improvement of $16.10 \%$.

We observe significant improvements across all the datasets with thought-based training approaches when compared to the label-based training approaches. This highlights that the learning for implicature recovery is also transferable to other datasets and pragmatic tasks. 

\begin{tcolorbox}[,
    colback=gray!10,     
    colframe=gray!30,    
    coltitle=black,
    fonttitle=\bfseries, 
    top=0mm,
    left=0mm,            
    right=0mm,           
    segmentation style={solid, black!80},
    bottom=-1mm-0.5mm,
    float=ht!
]
\small

\textbf{Given Input}:\\ 
Does Sentence 2 entail or contradict Sentence 1?\\
Sentence 1: The girl moved as fast as a turtle\\
Sentence 2: The girl was quick
\tcbline

\textbf{With Thought Prediction}:
Sentence 1 states that the girl moved at a speed similar to a turtle (slow), while Sentence 2 indicates that she was quick. These descriptions contradict each other.\\
Prediction: Contradiction

\textbf{Without Thought Prediction}: Entailment      \\
\textbf{Correct Label}: Contradiction

\tcbsubtitle[halign=center]{Illustration 2: General Prediction for NLI}

\end{tcolorbox} 
\label{box: transfer_learning_example}

\begin{table*}[htbp]
\centering

\resizebox{\textwidth}{!}{

\begin{tabular}{|l|ccc|ccc|ccc|c|}

\hline
\textbf{Setting}         & \multicolumn{3}{c|}{\textbf{FigQA (\%)}} & \multicolumn{3}{c|}{\textbf{Flute (\%)}} & \multicolumn{3}{c|}{\textbf{IMPPRES (\%)}} & \textbf{Mean F1 (\%)} \\
\hline
 & \textbf{P} & \textbf{R} & \textbf{F1} & \textbf{P} & \textbf{R} & \textbf{F1} & \textbf{P} & \textbf{R} & \textbf{F1} &  \\
\hline



\hline
\textbf{Zero-Shot}        &  62.14  &  62.13  &  62.13 & 77.77 &   77.77  &  75.70  &  43.16  &  37.62  &  39.19  & 59.00 \\ \hline
\textbf{DPO }             & 61.77 & 61.69 & 61.63 & 76.86 & 75.55 & 72.56 & 43.22 & 37.55 & 39.03 & 57.74 \\
\textbf{DPO + Thought} & 63.21 & 62.98 & 62.80 & 72.20 & 72.74 & 71.62 & 44.47 & 41.06 & 42.08 & 58.83 \textcolor{teal}{(+1.09)}  \\
\hline
\textbf{SFT }             & 59.59 & 58.99 & 58.35 & 76.28 & 73.35 & 69.40 & 49.37 & 48.75 & 44.27 & 57.34 \\
\textbf{SFT + Thought} & \textbf{64.09} & \textbf{63.85} &\textbf{ 63.69} & \textbf{78.13} & \textbf{78.30} & \textbf{76.36} & \textbf{49.72} & \textbf{49.48} & \textbf{46.72} & \textbf{62.25} \textcolor{teal}{(+4.91)}  \\ \hline


\end{tabular}
}

\caption{\textbf{Transfer Learning for NLI} on figurative sentences: FigQA, Flute, and IMPPRES.}
\label{tab: transfer_learning_nli}
\end{table*}

\begin{table*}[htbp]
\centering
\resizebox{\textwidth}{!}{
\begin{tabular}{|l|ccc|ccc|ccc|c|}
\hline
\textbf{Setting} & \multicolumn{3}{c|}{\textbf{Ludwig (\%)}} & \multicolumn{3}{c|}{\textbf{Presupposition (\%)}} & \multicolumn{3}{c|} {\textbf{Reference (\%)}} & \textbf{Mean F1 (\%)} \\
\hline
 & \textbf{P} & \textbf{R} & \textbf{F1} & \textbf{P} & \textbf{R} & \textbf{F1} & \textbf{P} & \textbf{R} & \textbf{F1} & \\
\hline
\textbf{Zero-Shot} & 70.92 & 65.21 & 61.42 & 52.10 & 50.55 & 17.41 & 11.31 & 26.33 & 15.82 & 31.55\\
\hline
\textbf{DPO} & 73.35 & 70.29 & 69.99 & 53.89 & 53.19 & 26.02 & 22.58 & 35.62 & 22.87  & 39.62\\
\textbf{DPO+Thought} & 74.07 & 72.71 & 72.75 & \textbf{53.36} & \textbf{55.77} & \textbf{51.51} & 68.41 & 62.64 & 62.88  & 62.38 \textcolor{teal}{(+22.76)}  \\ \hline
\textbf{SFT} & 61.68 & 60.96 & 59.73 & 52.47 & 50.92 & 19.02 & 3.27 & 6.30 & 4.31 & 27.68 \\
\textbf{SFT+Thought} & \textbf{76.42} & \textbf{76.53} & \textbf{76.40} & 54.23 & 57.64 & 44.27 & \textbf{69.08} & \textbf{69.77} & \textbf{69.34}  & \textbf{63.33} \textcolor{teal}{(+35.65)}\\ \hline

\end{tabular}
}
\caption{\textbf{Transfer Learning} for the Presupposition, Ludwig and Reference dataset}
\label{tab: transfer_learning_others}
\end{table*}

In the 
\hyperref[box: transfer_learning_example]{Illustration 2}, we describe a general scenario where the model is able to resolve the figurative language of \textit{as fast as a turtle} to \textit{slow}, finally arriving at the correct prediction of Contradiction.

\subsection{Data Ablations}
\label{data_ablation}
In this section, we discuss the effects of introducing our synthetically created data in the training. Specifically, we perform the experiments without training with any synthetic data and use the Circa data (similar to the original setting) to train our best-performing model, Gemma, in all four settings. We report the results in Table \ref{tab:data ablation} and note that the performances across all settings are significantly lower compared to the original training setup (as reported in Table \ref{tab:QA_data_comparison}), where we also include the synthetic data. We observe a reduction of 9.11\% in the SFT+Thought setting and 15.84\% in the DPO setting, respectively. A drastic reduction is observed in the DPO+Thought and SFT settings, primarily due to the poor legibility of model predictions where the generated text has very high hallucination. This shows the utility of synthetic data, highlighting its role in enhancing the robustness of training mechanisms by providing diverse and well-aligned examples that may be difficult to cover otherwise.

\subsection{Error Analysis}
This section describes the most prominent cases of error in the given task. We observe that the majority of errors occur when there is a use of a complex linguistic phenomenon that requires an additional layer of interpretation. 

\begin{tcolorbox}[,
    colback=gray!10,     
    colframe=gray!30,    
    coltitle=black,
    fonttitle=\bfseries, 
    top=0mm,
    left=0mm,            
    right=0mm,           
    segmentation style={solid, black!80},
    bottom=-1mm-0.5mm,
    float=ht!
]
\small

\textbf{Given Input}:\\
Does Sentence 2 entail or contradict Sentence 1?\\
Sentence 1: I was going to have dinner with friends today but I was stuck at work.\\
Sentence 2: I was so looking forward to dinner with friends today, but unfortunately, I got stuck at work.
\tcbline
\textbf{With Thought Prediction}:
The first sentence states that the person had plans for dinner with friends but couldn't go due to being stuck at work. The second sentence also mentions having dinner with friends but getting stuck at work, indicating similar circumstances. However, the first sentence uses "stuck," while the second uses "lovely." 

Prediction: Contradiction\\
\textbf{Correct Label}: Entailment
\tcbsubtitle[halign=center]{Illustration 3: Misinterpretation of sarcasm}
\end{tcolorbox} 
\label{box: error_case_sarcasm}

In the 
\hyperref[box: error_case_sarcasm]{Illustration 3}, we describe one such scenario. Specifically, sentence 2 shows the presence of sarcasm, which is evident in the latter part, which uses the adjective `lovely' for work. The model fails to understand that the speaker is being sarcastic about the situation, and the implication is similar to sentence 1, which leads to an incorrect prediction of `Contradiction'. Similar predictive behaviour is observed when other special linguistic phenomena like metaphors and hyperbole are present.





\subsection{Thought Analysis}
\label{subsec:thought_analysis}

In this section, we discuss various insights related to the thought generations. The primary aim of this study is to understand if the model is deriving the predictions from the correct thought or getting the predictions right with the incorrect logic. To analyze this quantitatively, we consider our best-performing model, Gemma2-2B-SFT, for evaluating thoughts using GPT4O-mini. To perform a human evaluation to ensure the quality of GPT4o-mini predictions, we asked two linguistic experts to annotate if they agreed with the predictions, leading to a Cohen kappa score of 0.79 on 85 examples randomly sampled from the data.

We observe that in the cases of correct predictions, the model generates correct thoughts in 96.41 \%  of the instances and 3.59 \% of incorrect thoughts. For incorrect predictions,  we observe 70.92\% instances with wrong thoughts while 29.08\% instances have the correct thoughts where the model generates the correct reasoning but could not resolve the correct predictions. We discuss an example in \hyperref[box: error_case_sarcasm]{Illustration 4} pertaining to the scenario where the model is correctly generating the contrast in both the sentences but provides prediction as `Entailment' instead of `Contradiction'.

In general, we observe that most of the correct predictions have correct thoughts, and a significant amount of incorrect predictions also have correct thoughts. This highlights that the model is able to generate thoughts to a reasonable extent but cannot cut the threshold of reaching the correct answer.

\begin{tcolorbox}[,
    colback=gray!10,     
    colframe=gray!30,    
    coltitle=black,
    fonttitle=\bfseries, 
    top=0mm,
    left=0mm,            
    right=0mm,           
    segmentation style={solid, black!80},
    bottom=-2.2mm-0.5mm,
    float=ht!
]
\small

\textbf{Given Input}: Determine whether a given response to a question implies a "Yes" or "No".\\
Context: X and Y are colleagues leaving work on a Friday at the same time.\\
Question: Do you have any big plans for the weekend?\\
Response: I have a number of things to do this weekend
\tcbline
\textbf{With Thought Prediction}:
The response indicates that the person has several activities planned for the weekend, suggesting they likely have some significant commitments.

Prediction: No\\
\textbf{Correct Label}: Yes
\tcbsubtitle[halign=center]{Illustration 4: Correct Thought but Incorrect Prediction}
\label{box:correct_thought_incorrect_prediction}
\end{tcolorbox}

\subsection{Thought Perturbation Analysis}
In this section, we describe our experiments, which aim to understand the if the improvements are observed due to the presence of correct thoughts leading to the right label in the training data 
or is it just some spurious correlation. For this experiment, we perturb the correct thought with the incorrect thought: for SFT, we replace the correct thought with the incorrect thought and for DPO, we flip the correct (preferred) and incorrect (dispreferred) thoughts in preference data. In general, we observe a significant decrement in the scores compared to the original setting where we consider the correct thought. The decrease in the SFT is very drastic, and the F1-scores went down to as low as $2\%$.  
In the DPO, we also see a considerable decline in the performance ($25\%-30\%$) across all QA tasks compared to the DPO+Thought settings. Even though, in both cases, there is a decrease in the performance, DPO models did not suffer a tragic decline in the accuracies due to the presence of the regularizing constant ($\beta$) in DPO. 
In other words, the regularizing constant $\beta$ prevents the large updates in the model, which is not the case in SFT, where the weight updates are unconstrained. 


\section{Conclusion and Future Work}
In this work, we highlighted the effectiveness of integrating explicit thought processes into two training paradigms: 1) Supervised Fine-Tuning (SFT) and 2) Direct Preference Optimization (DPO). Our findings indicate that while thought integration benefits both training approaches, thought-based SFT consistently outperforms its DPO counterpart in pragmatic reasoning tasks. Through a detailed analysis of model predictions, we uncover key patterns in implicature resolution and identify specific failure cases that illuminate areas for further improvement. To reinforce the role of structured reasoning, we investigate the impact of perturbing thought generation, revealing a notable decline in performance when the reasoning process is disrupted. Furthermore, our transfer learning experiments demonstrate the adaptability of thought-based training, showing its efficacy in generalizing across previously unseen datasets and pragmatic tasks. In the future, our goal is to refine this approach by developing a process-based reward mechanism that better aligns LLMs with human pragmatic inference, ultimately bridging the gap between computational and human-like language understanding.

\section{Limitations}
While our approach improves implicature recovery, it also presents several limitations. The reliance on explicit thought generation may introduce biases, as the quality and accuracy of generated thoughts depend on both the model's prior knowledge and human annotations. Additionally, the increased computational complexity associated with training thought-based models may limit scalability in resource-constrained settings. Further, while transfer learning results are promising, the generalizability of our approach across diverse linguistic domains, including highly contextual or culturally specific implicatures, remains an open question. In Future, we plan to explore more efficient training strategies and broader evaluation frameworks to enhance the robustness and applicability of thought-based learning.




\bibliography{acl_latex}

\appendix

\section{Evaluation of GPT-Generated Thoughts}
\label{appendix: thought_evaluation}

To evaluate the quality of the generated thoughts, we sampled 500 data points, including 200 from the CIRCA dataset, 150 from a synthetic CIRCA dataset, and 150 from a synthetic LUDWIG dataset. These data points were annotated by an external annotator and one of the authors of this paper. The evaluation focused on assessing the alignment between the correct label and its corresponding generated thought (Correct Label Thoughts - CLT), as well as the wrong label and its corresponding generated thought (Wrong Label Thoughts - WLT). Additionally, the confidence of alignment was rated on a three-point scale: 1 (Poor), 2 (Average), and 3 (Good). 

The results indicate that the correct label and its corresponding thought aligned 99\% of the time, and similarly, the wrong label and its corresponding thought also aligned 99\% of the time. Furthermore, both annotators agreed on the alignment 99\% of the time and showed 97\% agreement in confidence ratings. The external annotator was compensated at a standard rate for the annotation task. Table~\ref{tab:agreement} presents the agreement percentages between the two annotators for alignment and confidence evaluations.
\begin{table}[ht!]
    \centering
    \begin{tabular}{|l|c|}
        \hline
        \textbf{Evaluation Metric} & \textbf{Agreement Percentage} \\
        \hline
        CLT Alignment & 99.80\% \\
        CLT Confidence & 97.24\% \\
        WLT Alignment  & 100.00\% \\
        WLT Confidence & 98.82\% \\
        \hline
    \end{tabular}
    \caption{Agreement percentages between the two annotators for alignment and confidence evaluations.}
    \label{tab:agreement}
\end{table}

\section{Prompts and Templates Used for Generating Thoughts}
\label{appendix: generation_prompt}

In this section, we provide the prompts used for generating thoughts corresponding to the correct labels using GPT-4o mini, as well as the structured templates employed for generating thoughts related to the wrong labels.

\begin{tcolorbox}[,
    colback=gray!10,     
    colframe=gray!30,    
    coltitle=black,
    fonttitle=\bfseries, 
    top=0mm,
    left=0mm,            
    right=0mm,           
    segmentation style={solid, black!80},
    bottom=-1mm-0.5mm,
    float=ht!
]
\small

\textbf{Prompt}: Generate a one line rational to support [label] label to the answer-Y:\\
context        :   [context] \\
question-X     :   [question]\\
answer-Y       :   [answer]\\

\tcbsubtitle[halign=center]{Prompt For Generating Correct Label Thoughts}

\end{tcolorbox} 
\label{box: Prompt For Generating Correct Label Thoughts }

\begin{tcolorbox}[,
    colback=gray!10,     
    colframe=gray!30,    
    coltitle=black,
    fonttitle=\bfseries, 
    top=0mm,
    left=0mm,            
    right=0mm,           
    segmentation style={solid, black!80},
    bottom=-1mm-0.5mm,
    float=ht!
]
\small

\textbf{Example Templates for label “Yes”}: \\
1) Y’s response directly states their interest or intent, making the affirmative answer obvious.\\
2) Y explicitly agrees with the question, providing an unambiguous 'yes.'\\
3) The reply given by Y is positive and directly answers the question, ensuring clarity in the response.\\

\tcbline

\textbf{Example Templates for label “No”}:
\\
1) Y's response touches on related information but does not directly affirm the question, suggesting the answer may be 'no.'\\
2) While Y provides some context, the core question remains unanswered, implying that the response could be interpreted as a 'no.'\\
3) Y offers additional information but avoids directly addressing the question, indicating an implicit negative response.\\

\tcbsubtitle[halign=center] {Templates for Generating Wrong Label Thoughts}

\end{tcolorbox} 
\label{box:Templates for Generating Wrong Label Thoughts }

\section{Prompts Used for Training and Evaluation}
\label{Appendix: Train prompts}

In this section, we provide the prompts used for training the models using Supervised Fine-Tuning (SFT) with label tokens and SFT with thoughts.
We first present a generic prompt that serves as a foundational structure for training. This prompt is then adapted to fit the specific chat templates of each model, ensuring compatibility with their respective architectures and tokenization formats. The modifications may include adjustments in prompt wording, system instructions, formatting, or token placement to optimize the model’s performance across different setups.

\begin{tcolorbox}[,
    colback=gray!10,     
    colframe=gray!30,    
    coltitle=black,
    fonttitle=\bfseries, 
    top=0mm,
    left=0mm,            
    right=0mm,           
    segmentation style={solid, black!80},
    bottom=-1mm-0.5mm,
    float=ht!
]
\small
\textbf{Prompt for Training Data With Context
}: \\

\textbf{QA Context Input}: \\
You are reasoning driven assistant.\\
Given the following \textbf{context, question} and \textbf{response}, you task is to determine whether the response to a question implies a "Yes" or "No." Focus on the meaning implied in the response. \\
\textbf{Pretext} : [pretext] \\
\textbf{Question}: Does the response imply a "Yes" or a "No"? Do not output anything other than "Yes" or "No".\\

\textbf{QA context Output}: \\
prediction: [label]

\tcbline

\textbf{Prompt for Training Data Without Context:} \\

\textbf{QA Input}: \\
You are reasoning driven assistant.\\
Given the following \textbf{question} and a \textbf{response}, you task is to determine whether the response to a question implies a "Yes" or "No". Focus on the meaning implied in the response. \\
\textbf{Pretext} : [pretext] \\
\textbf{Question}: Does the response imply a "Yes" or a "No"? Do not output anything other than "Yes" or "No".\\

\textbf{QA Output}: \\
prediction: [label]

\tcbsubtitle[halign=center] {Prompt for Training QA Data with Label Tokens}

\end{tcolorbox} 
\label{box:Prompt for Training QA Data with and Without Context }

\begin{tcolorbox}[,
    colback=gray!10,     
    colframe=gray!30,    
    coltitle=black,
    fonttitle=\bfseries, 
    top=0mm,
    left=0mm,            
    right=0mm,           
    segmentation style={solid, black!80},
    bottom=-1mm-0.5mm,
    float=ht!
]
\small
\textbf{Prompt for Evaluating NLI
}: \\

\textbf{NLI Input}: \\
You are reasoning driven assistant.\\
Your task is to analyze the relationship between two sentences by first providing an explanation. Use the explanation to derive the prediction, which can be either "Entailment" or "Contradiction".\\
\textbf{Pretext} : [pretext] \\
\textbf{Question}: Analyze the relationship between the two sentences below and provide an explanation of your reasoning process. Derive the final prediction based on your explanation and give it in the below format:\\

\textbf{NLI Output}: \\
Explanation: [rationale], \\ 
prediction: [label]

\tcbsubtitle[halign=center] {Prompt for evaluating NLI tasks}

\end{tcolorbox} 
\label{box:Prompt for Evaluations }

\begin{tcolorbox}[,
    colback=gray!10,     
    colframe=gray!30,    
    coltitle=black,
    fonttitle=\bfseries, 
    top=0mm,
    left=0mm,            
    right=0mm,           
    segmentation style={solid, black!80},
    bottom=-1mm-0.5mm,
    float=ht!
]
\small
\textbf{Prompt for Training Data to Generate Thought
}: \\

\textbf{QA Context Thought Input}: \\
You are reasoning driven assistant.\\
Your task is to analyze whether a given response to a question implies a "Yes" or "No" by providing a one-line thought. The thought should focus on the reasoning process and should not include the final prediction.\\
\textbf{Input:} [pretext] \\
\textbf{Task:}  
Analyze the given context, question, and response. Provide a one-line reasoning thought without deriving the final prediction. Use the format below: \\  
\textbf{Thought:} " " \\

\textbf{QA context Thought Output}: \\
Thought: [rationale]

\tcbline

\textbf{Prompt for Training Data to Generate both Thought and Label} \\

\textbf{QA Context Thought Input}: \\
You are reasoning driven assistant.\\
Your task is to determine whether a given response to a question implies a "Yes" or "No" by first providing a one-line explanation. Use the explanation to derive the prediction, which can be either "Yes" or "No”
\textbf{Input}:  [pretext]

\textbf{Task}:  
Analyze the given context, question, and response. Provide a one-line of your reasoning process. Use the explanation to derive the prediction: "Yes" or "No" and give in the below format:\\
\textbf{Explanation:} "" \\
\textbf{Prediction:} ""  \\

\textbf{QA Context Thought Output}: \\
Explanation: [rationale],  \\
Prediction: [label]

\tcbsubtitle[halign=center] {Prompt for Training QA Data with Thoughts}

\end{tcolorbox} 
\label{box:Prompt for Training QA Data with Thoughts } 
Similarly, for the Direct Preference Optimization (DPO) task, we incorporate both the correct label thought and the wrong label thought during training. By including both perspectives, the model learns to distinguish between well-reasoned correct responses and incorrect alternatives, thereby improving its ability to align with human preferences. This approach enhances the model’s reasoning capabilities, ensuring that it not only recognizes correct answers but also understands why certain responses are less appropriate.

\section{Prompts Used for Evaluating Model Generated Thoughts}
\label{Appendix: Eval prompts}

In this section, we provide the prompts used for evaluating the thoughts generated by models which are trained using a thought-based training mechanism.

\begin{tcolorbox}[,
    colback=gray!10,     
    colframe=gray!30,    
    coltitle=black,
    fonttitle=\bfseries, 
    top=0mm,
    left=0mm,            
    right=0mm,           
    segmentation style={solid, black!80},
    bottom=-1mm-0.5mm,
    float=ht!
]
\small

\textbf{Prompt}: 
Your task is to verify whether the given sentences follow the ground truth. Only output yes or no. \\
\textbf{Ground Truth:} [Input] \\
\textbf{Reasoning:} [Correct Thought]\\
\textbf{Given Sentences:}
[model generated thought]

\tcbsubtitle[halign=center]{Prompt For Evaluating Model Generated Thoughts}

\end{tcolorbox} 
\label{box: Prompt For Generating Correct Label Thoughts }

\end{document}